\documentclass[lettersize,journal]{IEEEtran}
\usepackage{amsmath,amsfonts}
\usepackage{algorithmic}
\usepackage{algorithm}
\usepackage{array}
\usepackage[caption=false,font=normalsize,labelfont=sf,textfont=sf]{subfig}
\usepackage{textcomp}
\usepackage{stfloats}
\usepackage{url}
\usepackage{verbatim}
\usepackage{graphicx}
\usepackage{cite}
\usepackage{amssymb}
\usepackage{amsmath}
\usepackage{txfonts}
\usepackage{mathdots}
\usepackage{graphicx}
\usepackage{rotating}
\usepackage{caption}
\usepackage{subcaption}
\usepackage{lineno,hyperref}
\usepackage{changepage}
\usepackage{setspace}
\usepackage{multirow,algorithm}
\usepackage[algo2e]{algorithm2e}
\usepackage{enumerate}
\usepackage{cite}
\usepackage{tabularx,booktabs,caption,paralist}
\usepackage[table]{xcolor}
\usepackage{multirow,paralist}
\begin{document}
	
\begin{center}
	\parbox{\textwidth}{\centering \bfseries This work has been submitted to the IEEE for possible publication. Copyright may be transferred without notice, after which this version may no longer be accessible.}
\end{center}
\vspace{1cm} % Adjust spacing as needed

\title{Robust Energy Consumption Prediction with a Missing Value-Resilient Metaheuristic-based Neural Network in Mobile App Development }

\author{Seyed Jalaleddin Mousavirad and Luís A. Alexandre
        % <-this % stops a space
\thanks{Seyed Jalaleddin Mousavirad is with the Universidade da Beira Interior, Covilhã, Portugal. Luís A. Alexandre is with Universidade da Beira Interior and NOVA LINCS, Covilhã, Portugal.}% <-this % stops a space
\thanks{Manuscript received ...; revised ...}}

\markboth{Journal of \LaTeX\ Class Files,~Vol.~14, No.~8, August~2021}%
{Shell \MakeLowercase{\textit{et al.}}: A Sample Article Using IEEEtran.cls for IEEE Journals}

\IEEEpubid{0000--0000/00\$00.00~\copyright~2021 IEEE}

\maketitle

\begin{abstract}
Energy consumption is a fundamental concern in mobile application development, bearing substantial significance for both developers and end-users. 
%Moreover, it is a critical determinant in the consumer's decision-making process when considering a smartphone purchase. 
%From the sustainability perspective, it becomes imperative to explore approaches aimed at mitigating the energy consumption of mobile devices, given the significant global consequences arising from the extensive utilisation of billions of smartphones, which imparts a profound environmental impact.
%From a sustainability standpoint, it is crucial to find ways to reduce the energy use of mobile devices. The widespread use of smartphones worldwide has a major environmental impact, highlighting the need for energy-efficient solutions.}
 %Despite the existence of various energy-efficient programming practices within the Android platform, the dominant mobile ecosystem, there remains a need for documented machine learning-based energy prediction algorithms tailored explicitly for mobile app development.
  Main objective of this research is to propose a novel neural network-based framework, enhanced by a metaheuristic approach, to achieve robust energy prediction in the context of mobile app development. The metaheuristic approach here aims to achieve two goals: 1) identifying suitable learning algorithms and their corresponding hyperparameters, and 2) determining the optimal number of layers and neurons within each layer. 
  %The metaheuristic approach here plays a crucial role in not only identifying suitable learning algorithms and their corresponding parameters but also determining the optimal number of layers and neurons within each layer.
  %To the best of our knowledge, prior studies have yet to employ any metaheuristic algorithm to address all these hyperparameters simultaneously. 
  Moreover, due to limitations in accessing certain aspects of a mobile phone, there might be missing data in the data set, and the proposed framework can handle this. In addition, we conducted an optimal algorithm selection strategy, employing 13 base and advanced metaheuristic algorithms, to identify the best algorithm based on accuracy and resistance to missing values.
  	The representation in our proposed metaheuristic algorithm is variable-size, meaning that the length of the candidate solutions changes over time. We compared the algorithms based on the architecture found by each algorithm at different levels of missing values, accuracy, F-measure, and stability analysis. Additionally, we conducted a Wilcoxon signed-rank test for statistical comparison of the results. The extensive experiments show that our proposed approach significantly improves energy consumption prediction. 
  	Particularly, the JADE algorithm, a variant of Differential Evolution (DE), DE, and the Covariance Matrix Adaptation Evolution Strategy deliver superior results under various conditions and across different missing value levels.
    
\end{abstract}

\begin{IEEEkeywords}
Energy Consumption Prediction, Mobile Application Development, Evolutionary Computation, Neural Network, hyperparameters, Optimisation
\end{IEEEkeywords}

\section{Introduction}
\label{sec:Intro}
\IEEEPARstart{I}{n} contemporary mobile application development, developers express profound concerns regarding the impact of their Apps on the battery life of smartphones. Many apps are harassed in App stores because of their tendency to consume too much power~\cite{smartphone_03}. Consequently, developers diligently acknowledge this energy issue and actively seek assistance to address it, albeit finding satisfactory guidance remains a rarity~\cite{smartphone_06}. To promote enhanced energy efficiency, mobile device manufacturers offer comprehensive guidelines for developers. Moreover, the energy consumption factor significantly influences the overall satisfaction of mobile device consumers~\cite{smartphone_03}. 
%Notably, recent research has underscored the pivotal role of battery life as a decisive criterion shaping the smartphone purchasing decisions of 1898 surveyed users in the US~\cite{smartphone_03,smartphone_04}.
 This mounting concern over energy consumption has been corroborated by staggering statistics, revealing that a substantial majority of consumers, approximately 90\%, struggle with the phenomenon of low battery anxiety~\cite{smartphone_05}.

%Considering sustainability, finding ways to make mobile devices use less energy is crucial. The billions of phones we use today also significantly impact the world. Remarkably, the ecological impact of our digital device utilisation, encompassing smartphones and tablets, is projected to surpass that of the aviation industry in terms of its contribution to global warming~\cite{smartphone_07}. Therefore, prioritising energy efficiency in mobile devices is crucial in addressing environmental concerns and fostering sustainable practices.

Analysing and optimising energy consumption in mobile devices pose a challenging and time-consuming endeavour for both end-users and developers. Effectively monitoring an application's energy usage necessitates conducting numerous tests under diverse conditions and across multiple devices, which demands considerable time and effort. Developers frequently employ multiple monitoring tools, leading to contextually bound findings. Moreover, the Android platform demonstrates substantial diversity, accommodating a vast population of approximately 3.3 billion Android smartphone users in 190 countries globally.\footnote{Source: http://www.bankmycell.com/blog/how-many-android-users-are-there}. 
%However, developers' understanding of hardware performance remains confined to their devices. Without access to suitable tools and expertise, they encounter challenges in effectively comparing energy consumption patterns among different Apps or comprehending App behaviour across diverse devices and scenarios. Additionally, the energy dynamics of an App exhibit variations contingent on its usage, including factors such as operating system versions and the activation status of various hardware components. Such distinctions must be thoughtfully considered during comparative analyses to ensure accurate evaluations.

Over the past few years, numerous research studies~\cite{Energy_APP08, Energy_APP_09,Energy_APP_10,Energy_APP07,Energy_APP11,Energy_APP12} have explored energy-aware programming trends in the Android platform and sought to identify more efficient alternatives. However, automatic energy consumption prediction has received limited attention. For instance, \cite{smartphone_08} investigated the potential of proxy metrics like CPU usage and memory write operations as predictors for energy consumption in a music streaming application on mobile devices. Their experiments involved two Apps, namely Spotify music and podcast App, to understand the correlation between the selected proxies and energy consumption. \cite{Energy_APP05} investigates how machine learning (ML) techniques perform in predicting the energy consumption of Java classes within Android applications. The study focuses solely on utilising structural properties derived from source code analysis. Their initial findings indicate that learning models can not perform well in this application. \cite{Energy_APP06} suggested utilising byte-code transformations in conjunction with genetic search methods to decrease the energy usage of an application.

In another study~\cite{smartphone_09}, researchers conducted a large-scale user study to measure the energy consumption characteristics of 15500 BlackBerry smartphone users. They compiled a substantial data set to create the Energy Emulation Toolkit (EET), facilitating developers in comparing their Apps' energy requirements with real user energy traces. Additionally, \cite{smartphone_10} leveraged ML techniques and environment data to predict a smartphone's next unlock event. Through a 2-week field study involving 27 participants, they demonstrated the feasibility of accurately predicting unlock events, leading to energy savings by optimising software-related background processes. 
%They suggested reducing energy consumption using short-term predictions to avoid unnecessary executions or initiating resource-intensive tasks, such as OS updates, when the phone is locked. By predicting the next phone unlock event, smartphones can pre-collect sensor data or prepare content to enhance the user experience during subsequent usage. 
Moreover, a recent study by \cite{JPEG_Postdoc01} identified two image-based proxies for energy consumption: image file size and image quality. They found that increasing image file size and quality directly impacts energy consumption. To address this issue, they proposed a multi-objective approach to optimise image size and quality, while maintaining a small energy footprint. However, this approach needed to have consideration of user opinions. As a result, \cite{JPEG_Postdoc02} incorporated user preferences by altering the encoding representation to refine the energy consumption prediction process further. In another recent research, \cite{JPEG_Postdoc03} presents a machine-learning technique for predicting smartphone energy consumption. Their approach utilises the Histogram-based Gradient Boosting Classification Machine (HGBC) for the modelling phase. However, an essential aspect of smartphones, namely permissions, is overlooked in their algorithm. As a result, some users may only grant some of the required permissions to an App. Consequently, certain features may be absent from the feature set due to the unavailability of corresponding permissions.

This paper proposes a robust energy consumption prediction with a missing value-resilient metaheuristic-based artificial neural networks (ANNs) approach in mobile App development. ANNs are popular and influential machine learning models inspired by the human brain's structure and functioning. However, to achieve optimal performance with ANNs, it is essential to tune hyperparameters carefully. For ANNs, hyperparameters include the number of layers, the number of neurons in each layer, learning rate, solver, and momentum, among others. Tuning hyperparameters is critical because it significantly impacts the neural network's performance and generalisation ability. If the hyperparameters are not set appropriately, the network may suffer from overfitting, where it performs well on the training data but fails to generalise to new, unseen data. A comprehensive survey on hyperparameter optimisation is available in \cite{hyperparameter_optimization_survey}.
On the other hand, setting hyperparameters too conservatively may result in underfitting, where the network fails to capture the underlying patterns in the data. Also, finding the correct number of layers and neurons is essential for controlling the model's complexity. Deep neural networks with multiple layers and neurons can potentially learn intricate patterns in the data but also require more data and computational resources. Shallower networks may not capture complex relationships adequately. Therefore, finding the optimal balance is crucial for achieving the best performance. As a result, this paper proposes a novel missing value-resilient metaheuristic-based ANN for finding the optimal hyperparameters. 

More precisely, this paper presents several contributions, encompassing the following key aspects:
\begin{itemize}
	\item We introduce a missing value-resilient machine learning-based approach for predicting energy consumption in mobile App development. Notably, this work is the first study to address energy prediction in this domain while accounting for missing values in the data. 
%	\item Our machine learning algorithm leverages a tuned artificial neural network (ANN) during the modelling phase. By carefully adjusting the ANN's hyperparameters, we enhance its performance and predictive capabilities, ensuring more accurate and reliable energy consumption forecasts.
	\item We propose a general metaheuristic-based approach for hyperparameter tuning in ANNs. The utilisation of metaheuristic techniques facilitates an efficient search for optimal hyperparameter configurations, contributing to the overall effectiveness of our energy prediction model.
	\item As our proposed metaheuristic approach remains algorithm-independent, we applied our proposed method to 13 basic and advanced metaheuristic algorithms to identify the most robust and effective strategy for hyperparameter tuning. A pivotal goal of this paper involves benchmarking various metaheuristic algorithms for hyperparameter tuning in ANNs. Through rigorous evaluation, we aim to identify the most suitable and efficient metaheuristic technique for our energy prediction model.
	\item To the best of our knowledge, most of the employed metaheuristic algorithms have not been previously utilised in hyperparameter tuning for ANNs, nor in other contexts. Consequently, this paper represents the first endeavour to investigate the effectiveness of these algorithms in addressing this specific problem. 
\end{itemize}	

%In addition, from a sustainability perspective, our approach is significant in green computing. By optimising energy consumption predictions in mobile App development, this methodology can contribute to more energy-efficient practices, aligning with sustainable computing.
	
The remainder of this paper is organised as follows.  Our proposed algorithm is described in Section~\ref{sec:proposed}, while Section~\ref{sec:results} validates the proposed algorithm in the different conditions. Finally, we present the conclusions in Section~\ref{sec:conc}.

\section{Search Strategies}
For search strategies, there is flexibility in selecting various types of population-Based Metaheuristic (PBMH) techniques. However, it is impractical to analyse every PBMH technique available in the literature due to its vast and diverse collection. Therefore, we selected several well-established and state-of-the-art variants commonly utilised in evolutionary and swarm computing, leading to 13 algorithms. This wide range of choices not only aids in choosing the best algorithm for our search strategy but also can be used as a benchmark study.

\subsubsection{Base Algorithms} 

\begin{compactitem}
	\item
	\textbf{Genetic algorithm (GA)}~\cite{GA_Main_Ref}: The GA is the most established algorithm that consists of three key components: crossover, mutation, and selection. Crossover merges information from parent solutions, while mutation introduces random alterations to one or multiple elements of a candidate solution. The selection process in GA determines which solutions are carried forward to subsequent iterations based on the principle of favouring the most successful ("fittest") solutions. There are several ways for the selection process, while this paper focuses on tournament selection due to some advantages like less bias towards the fittest.
	\item 
	\textbf{Differential Evolution (DE)}~\cite{DE_Original}: DE employs three primary operators: mutation, crossover, and selection. The mutation operator generates candidate solutions by calculating a mutant vector using a scaled difference among candidate solutions. 
	%The most famous mutation is "DE/rand/1"., which represented as 
	%\begin{equation}
	%	v_{i}=x_{r1} + F (x_{r2}-x_{r3}) ,
	%\end{equation}
	%where $F$ denotes a scaling factor, and $x_{r1}$, $x_{r2}$, and $x_{r3}$ represent three distinct randomly chosen candidate solutions from the existing population. 
	The crossover operator combines the mutant vector with a target vector selected from the current population. Finally, the selection operator determines the candidate solution to be retained.
	\item \textbf{Memetic Algorithm (MA)}~\cite{Memetic_Main}: MA is a search strategy that incorporates both a population-based algorithm, such as GA and a local search approach. In the particular version we employed, each agent is assigned a probability indicating whether or not a local search operation should be performed.
	
	\item
	\textbf{Particle Swarm Optimisation (PSO)}~\cite{PSO_Main_Paper}: PSO is a PBMH inspired by swarm behaviour, where the updating process relies on the best position of each candidate solution and a global best position. The velocity vector of a particle is updated as
	\begin{equation}
		\label{Eq:PSO}
		v_{t+1}= \omega v_{t}+c_{1} r_{1} (p_{t}-x_{t})+c_{2} r_{2} (g_{t}-x_{t}) ,
	\end{equation} 
	
	where $t$ denotes the current iteration, $r_{1}$ and $r_{2}$ represent random numbers drawn from a uniform distribution in the range [0,1], $p_{t}$ corresponds to the personal best position, and $g_{t}$ represents the global best position. Also, $c_{1}$ and $c_{2}$ are called cognitive coefficient and social coefficient, respectively, while $\omega$ signifies inertia weight. 
	\item 
	\textbf{Covariance Matrix Adaptation Evolution Strategy (CMA-ES)}~\cite{CMA_main_paper}: CMA-ES algorithm is another PBMH employed in this paper that belongs to the class of evolution strategies. It utilises a probabilistic model to adaptively update the search distribution and guide the search towards promising regions of the search space. At each iteration, CMA-ES maintains a population of candidate solutions represented as a set of multivariate Gaussian-distributed samples. The algorithm dynamically adjusts this distribution's mean vector and covariance matrix based on the success or failure of the sample solutions.
	% The fitness evaluations of the candidate solutions guide the update of the mean vector. The algorithm calculates the weighted average of the successful solutions to obtain a new mean vector estimate. This allows CMA-ES to explore regions of the search space with higher fitness values. To update the covariance matrix, CMA-ES employs a rank-one update mechanism. It calculates the outer product of the mean-centred successful solutions, which is then used to update the covariance matrix estimate. This update ensures that the covariance matrix aligns with the successful search directions and adapts to the underlying problem landscape.
\end{compactitem}
\subsubsection{Advanced Algorithms} 
\begin{compactitem}
	\item
	\textbf{Self-organising Hierarchical PSO with Jumping Time-varying Acceleration Coefficients (HPSO)}~\cite{HPSO-TVAC}: The key features of HPSO can be summarised as follows:
	
	\begin{enumerate}
		\item Mutation is incorporated into the PSO algorithm.
		\item A new approach, the self-organising hierarchical particle swarm optimiser with TVAC, is introduced. It focuses on the social and cognitive aspects of the particle swarm strategy when determining each particle's updated velocity. Additionally, particles are reset to their initial state if they become stagnant in the search space.
		\item The PSO algorithm includes a time-varying mutation step size.
	\end{enumerate}
	\item 
	\textbf{Chaos PSO (CPSO)}~\cite{Chaos-PSO}: CPSP incorporates two techniques to enhance its performance: an adaptive inertia weight factor (AIWF) and a chaotic local search (CLS). The AIWF dynamically adjusts the inertia weight, denoted as $\omega$, in the original PSO algorithm based on the objective function value as
	\begin{equation}
		\omega=\begin{cases}
			\omega_{min}+\frac{(\omega_{max}-\omega_{min})(f-f_{min})}{f_{avg}-f_{min}} & f \leq f_{avg} \\
			\omega_{max} & f \geq f_{avg}
		\end{cases} ,
	\end{equation}
	where $\omega_{\text{max}}$ and $\omega_{\text{min}}$ represent the maximum and minimum values of $\omega$, respectively. The current objective function value of a candidate solution is denoted as $f$, while $f_{\text{avg}}$ and $f_{\text{min}}$ represent the average and minimum values of all candidate solutions, respectively.
	
	To further improve the effectiveness of CPSO, the CLS operator is also employed as a local search mechanism around the best position.
	% The CLS equation is as follows:
	%\begin{equation}
	%	\label{CLS}
	%	cx_{i}^{k+1}=4cx_{i}^{k}(1-cx_{i})
	%\end{equation}
	%where $cx_{i}$ represents the $i$-th chaotic variable, while $k$ denotes the iteration number. The chaotic variable $cx_{i}$ is distributed between 0 and 1. The equation exhibits chaotic behaviour when the initial value $cx_{0}$ lies in the range (0, 1), and the corresponding value $x_{0}$ is not equal to 0.25, 0.5, or 0.75.
	\item
	\textbf{Comprehensive Learning PSO (CLPSO)}~\cite{CLPSO02}: To prevent premature convergence, a comprehensive learning (CL) strategy is proposed for particle learning. Instead of relying solely on their own best positions (pbest), all particles' pbest can be utilised to adjust the velocity of each particle. The updating scheme in CLPSO is defined as follows:
	\begin{equation}
		\label{Eq:vel2}
		v_{t+1}^{i}= \omega v_{t}^{i}+c_{1}  r (pbest_{fi(d)}^{f}-x_{t}^{i}),
	\end{equation}
	where $fi(d)$ determines which particles' \textit{pbest} a specific particle $i$ should follow. The decision to learn from nearby particles is based on the comprehensive learning probability, $PC$. For each dimension, a random number is generated from a uniform distribution. If the generated random number exceeds $PC(i)$, the particle updates based on its pbest. Otherwise, it updates by incorporating information from nearby particles.
	\item 
	\textbf{DE with Self-Adaptation Populations (SAP-DE)}~\cite{SAP_DE}: It introduces self-adaptive features for population size, crossover rates, and mutation rates. Two variants, $SAP-DE-ABS$ and $SAP-DE-REL$, are proposed to determine the population size parameter $\pi$.
	
	In $SAP-DE-ABS$, $\pi$ is calculated by rounding the sum of an initial population size $NP_{ini}$ and a random value drawn from a standard normal distribution.
	On the other hand, $SAP-DE-REL$ initialises the population size parameter by sampling a value from a uniform distribution ranging from -0.5 to +0.5.
	% During each stage, the $\pi$ parameter needs to be updated. In $SAP-DE-REL$, the current population size is adjusted based on a growth rate, either increasing or decreasing by a certain percentage. In $SAP-DE-ABS$, the population size for subsequent generations is determined as the average population size attribute from all individuals in the current population.
	%	\textbf{ Self-Adaptive Differential Evolution (SADE)}~\cite{SADE_DE}: SADE is an enhanced version of DE that incorporates two mutation operators, namely \textit{DE/rand/1} and \textit{DE/current-to-best/1}, simultaneously. The \textit{DE/current-to-best/1} operator is defined as follows:
	%		\begin{equation}
		%		\label{SADE}
		%		v_{i}=x_{i}+ F_{i} . (x_{best}-x_{i})+ F_{i} . (x_{r1}-x_{r2}),
		%	\end{equation}
	%	where $x_{best}$ represents the best candidate solution from the current population, while $x_{r1}$ and $x_{r2}$ are two randomly selected candidate solutions. The scaling factor $F_{i}$ is applied to the $i$-th candidate solution.
	\item 
	\textbf{Adaptive Differential Evolution with Optional External Archive (JADE)}~\cite{DE_Jade01}: JADE adjusts the likelihood of generating offspring by either mutation operators based on the success ratio over the previous 50 generations, leading to select, gradually, the best mutation strategy for a specific problem.  Also, JADE employs a normal distribution for setting the scale factor for each candidate solution, leading to maintaining both local (with small $F_{i}$ values) and global (with large $F_{i}$ values) capabilities simultaneously.
	\item 
	\textbf{Success-History-based Parameter Adaptation for Differential Evolution (SHADE)}~\cite{SHADE01}:
	% The SHADE algorithm is a variant of DE that incorporates adaptive mechanisms to enhance its performance in solving optimisation problems.
	% It focuses on adapting the control parameters of DE during the evolution process. 
	The main idea behind SHADE is to maintain a success-history archive that stores information about previous successful solutions. 
	%This archive guides the search towards promising regions of the search space.
	In the archive updating strategy, SHADE uses a memory-based mechanism to update the success-history archive. Only the best solutions, which outperform their parents, are considered for archive inclusion.
	% This ensures that the archive contains high-quality solutions that have exhibited better performance. 
	Another characteristic of SHADE is the parameter adaptation strategy that aims to dynamically adjust the control parameters of DE, such as the crossover rate and the mutation factor, based on the information stored in the success-history archive. 
	%By considering the performance of previously successful solutions, SHADE adapts these parameters to suit better the characteristics of the optimisation problem being solved.
	\item 
	\textbf{SHADE using Linear Population Size Reduction (LSHADE)}~\cite{LSHADE01}: LSHADE is an extension of the SHADE algorithm, which aims to further improve the performance of SHADE by introducing two key components: the population size reduction factor and the memory size factor. The population size reduction factor determines the rate at which the population size is decreased over iterations. By gradually reducing the population size, LSHADE allocates more computational resources to the most promising solutions, thereby enhancing search efficiency. %On the other hand, the memory size factor determines the size of the success-history archive used in LSHADE. This archive stores information about previous successful solutions and guides the search towards promising regions of the search space. By adjusting the memory size factor, LSHADE controls the balance between exploration and exploitation during optimisation.
	\item
	\textbf{Phasor PSO(PPSO)}~\cite{Phasor-PSO}: PPSO suggests using control parameters based on a mathematical concept known as the phase angle ($\theta$). The velocity update in each iteration is determined as
	
\begin{multline}
	v_{i}^{iter} = |cos\theta_{i}^{iter}|^{2sin\theta_{i}^{iter}} \times (Pbest_{i}^{iter}-x_{i}^{iter}) \\
	\quad + |sin\theta_{i}^{iter}|^{2cos\theta_{i}^{iter}} \times (Gbest_{i}^{iter}-x_{i}^{iter})
\end{multline}

	where the velocity of each particle is influenced by the values of the phase angle ($\theta$) and is adjusted based on the differences between the personal best ($Pbest$), global best ($Gbest$), and the current position ($x$) of the particle.
	
\end{compactitem}

%\begin{algorithm2e}[tb]
%	\scriptsize
%	\SetAlgoLined
%	\SetKwInOut{Input}{Input}\SetKwInOut{Output}{Output}
%	\Input{ $D$: dimensionality of problem; $NFE_{\max}$: maximum number of function evaluations; $NP$: population size}
%	\Output{ $x^*$: the best individual }
%	\BlankLine
%    Split the dataset into two parts: train ($TrainSet$) and test ($TestSet$) sets \;
%    $t\leftarrow 0$ \;
%	\For{$i\leftarrow 1$ \KwTo $MaxHiddenLayer$} {
	%		Randomly generate initial population $Pop_{0}$\ based on the proposed representation and $i$ \;
	%	    $model_{best} \leftarrow$ the best candidate solution from the current population \;
	%		$NFE \leftarrow NP$ \;
	%		\While{$NFE<=NFE_{max}$} {
		%		$t\leftarrow t+1$ \;	
		%		$pop_{t} \leftarrow$ Apply the operators in the search strategy to generate a new population \;
		%		Calculate the objective function for each new candidate solution on the training set \;
		%		$model_{i} \leftarrow$ the best candidate solution from the current population \;
		%		\If{$f(model_{i}) < f(model_{best}) $} {
			%			  $model_{best} \leftarrow  model_{i}$
			%			}
		%		}
	%	}
%	Evaluate the $model_{best}$ by using k-fold cross validation 

%	\caption{The top-level view of the proposed Neuroevolution algorithm in the form of Pseudo-code.}
%	\label{Alg1:proposed}
%\end{algorithm2e}

\section{The Proposed Algorithm }
\label{sec:proposed}
This research introduces a universal approach using metaheuristic techniques to discover optimal hyperparameters within MLNN for mobile app development. This methodology can be seamlessly combined with various optimisation algorithms. The process involves integrating our suggested method into 13 distinct algorithms. Initially, we outline the data set we employ, then explain how we represent solutions and define the objective function. Subsequently, by incorporating the proposed approach into 13 foundational optimisation algorithms, we generate 13 new and distinct algorithms.

\subsection{General Structure}

In this work, we introduce a neuroevolution methodology aimed at predicting energy consumption within mobile application development. Our machine learning-based energy prediction approach offers can be deployed within Android app development through various techniques as follows.
	\begin{itemize}
		\item Embedded-App Technique: In this approach, the machine learning component is integrated into individual apps. However, this may lead to redundancy as multiple apps on the same device may have their own instance of the ML component.
		\item Embedded-Android Technique: Alternatively, the machine learning component could be integrated into the Android operating system itself as a built-in feature. While this approach may offer optimal performance, its feasibility depends on the decision of major stakeholders such as Google, which currently seems unlikely.
		\item Independent-App Technique: In this method, the machine learning component is packaged as a standalone app. However, convincing users to install an additional app alongside their desired applications may pose a challenge.
		\item Web Service Technique: Another option is to deploy the machine learning component as a web service. In this scenario, when an application requires energy consumption prediction, it can send a request to the web service and receive a recommendation. This approach remains effective until major Android stakeholders deem a built-in machine learning-based energy prediction component necessary.
	\end{itemize}		
The cornerstone of this methodology is its capability to dynamically tailor the architecture of MLNNs across a broad spectrum of optimisation algorithms. This adaptability is facilitated through the employment of thirteen distinct search strategies, thereby generating a diverse suite of algorithms specifically designed for predicting energy consumption in mobile app development. The algorithms we developed for this prediction include GA-AMLNN, DE-AMLNN, MA-AMLNN, PSO-AMLNN, CMA-ES-AMLNN, HPSO-AMLNN, CPSO-AMLNN, CLPSO-AMLNN, SAPE-DE-AMLNN, JADE-AMLNN, SHADE-AMLNN, LSHADE-AMLNN, and PPSO-AMLNN, all of which are based on PBMH algorithms. For example, LSHADE-AMLNN utilises the LSHADE algorithm to optimise the selection of hyperparameters, as well as the number and configuration of neurons and layers in the neural networks.
 
A pivotal element of our proposed framework is the integration of a variable, denoted as \textit{i}, which governs the number of hidden layers within the MLNNs. This variable not only specifies the initial configuration but also delineates the maximum allowable layers, directly influencing the dimensional complexity of each candidate solution. For illustrative purposes, an \textit{i} value of 4 implies the construction of a candidate solution with four hidden layers.

At the inception of our methodology, a diverse array of MLNNs, each with varying hyperparameters and neuron counts, is generated. This initial population is derived using PBMH algorithms, signifying a wide exploratory scope from the outset. The search strategies are then tasked with refining these architectures, guided by a specifically defined objective function aimed at identifying the most effective network architecture for energy consumption prediction. 

During the evolutionary phase, the search strategies increment the \textit{i} variable following the evaluation of each layer's efficacy, thereby methodically enhancing the network's complexity within predefined bounds. The evaluation of the evolved MLNN architectures is conducted through a rigorous 10-fold cross-validation process. This entails partitioning the original dataset into distinct training and testing subsets, with the training subset further subdivided to facilitate the validation process. This methodology not only underscores the robustness of our approach but also ensures its applicability and relevance to the dynamic field of mobile application development.

\begin{figure*}[!htb]
	\centering
	\includegraphics[width=1.7\columnwidth]{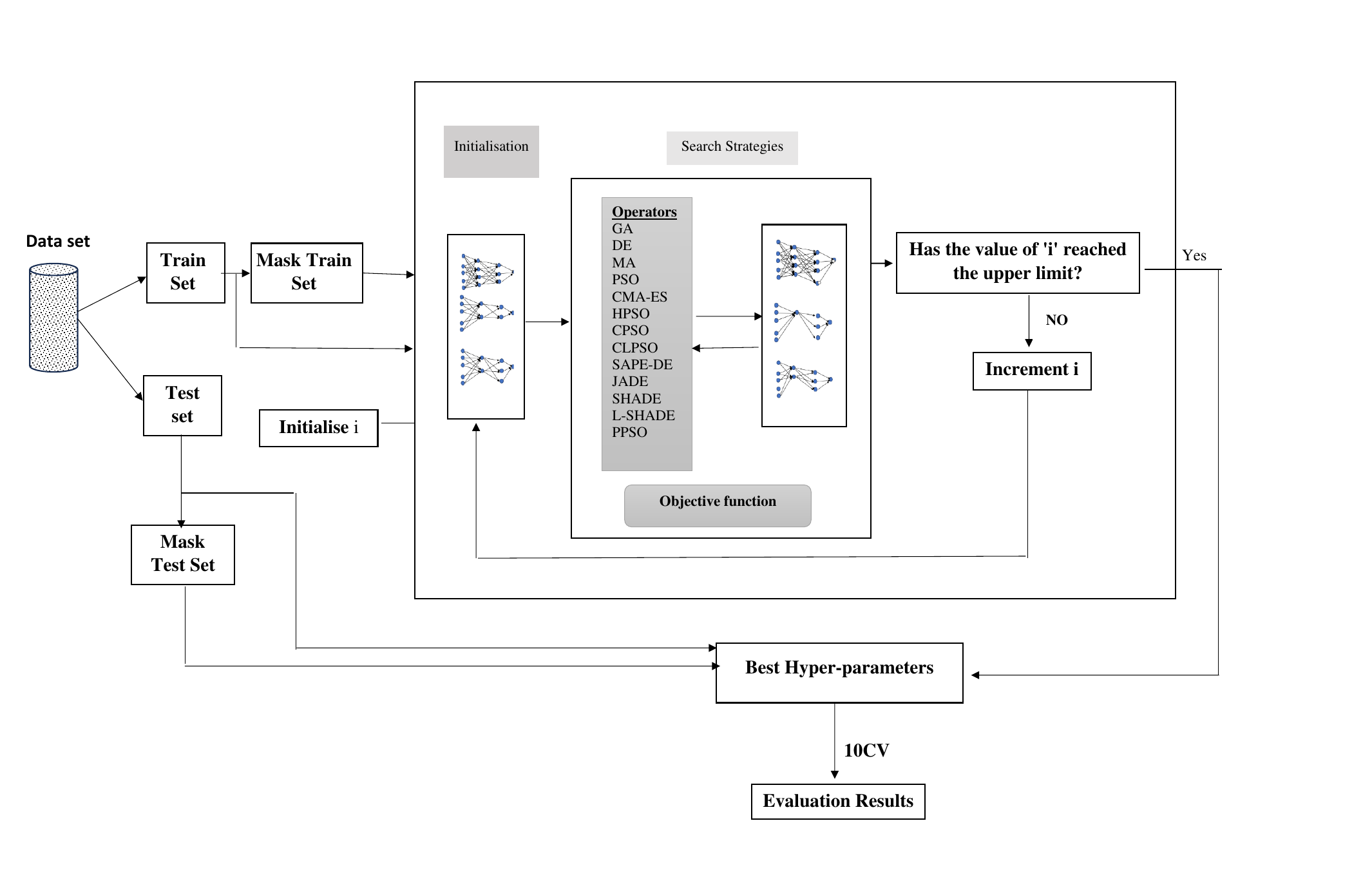}
	\caption{The general framework of our proposed algorithm.} 
	\label{fig:framwork}
\end{figure*}

\subsection{Preprocessing}

This particular step in our study involves transforming and encoding the existing data set to ensure compatibility with machine learning algorithms. First, we systematically removed instances labelled as "battery-state=charging" from the data set, as our primary objective is to predict energy consumption, specifically in the discharging state. In addition, we introduced a novel metric, Energy Consumption Per Minute (ECPM), designed to quantify the energy consumed per minute by a smartphone. The ECPM is defined as

\begin{equation}
	ECPM = \frac{BT_{state_{1}} - BT_{state_{2}}}{TS_{state_{1}} - TS_{state_{2}}} \times 60
\end{equation}

where $BT_{state_{i}}$ represents the battery level in state \textit{i}, and $TS_{state_{i}}$ refers to the timestamp in state \textit{i}. State$_{1}$ and State$_{2}$ are consecutive states, capturing the current settings of the smartphone, including features such as Bluetooth and Wifi states (on/off). The key assumption here is that the smartphone settings remain unchanged between consecutive states. ECPM is a valuable metric for evaluating energy consumption, with lower ECPM values indicating reduced energy usage. To ensure the metric's accuracy, instances where the 'battery-state=charging' state is positioned between two 'battery-state=discharging' states, along with the instance immediately following it, are excluded from the ECPM calculations.

\begin{figure}[tb]
	\centering
	\includegraphics[width=0.5\columnwidth]{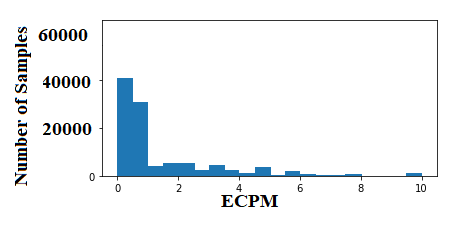}
	\caption{Histogram of randomly-selected instances } 
	\label{fig:histogram}
\end{figure}

As the present study focuses on energy prediction viewed as a classification problem, each instance has been categorised into one of three classes: safe, warning, or critical status, based on the Energy Consumption Per Minute (ECPM) metric. The assignment of each sample to a class is contingent upon the mobile App developer's discretion. For instance, an ECPM value of 0.5 might be considered high for one developer, while another could consider it safe. Since the developer's preferences are unknown, a histogram analysis has been employed to determine the class assignments. Figure~\ref{fig:histogram} shows the histogram for ECPM for 120000 randomly-selected instances. We tried to distribute the samples to different classes so that the classes remain almost balanced. However, it can be changed according to the developer's preference. According to the histogram, instances with ECPM values less than 0.5 are assigned to the first class. At the same time, those exceeding 1.5 are allocated to a separate class, with the remaining instances falling into the third class. 

\begin{comment}

\subsection{Proposed Neuroevolution Model }
This section explains our proposed neuroevolution model for predicting energy consumption in mobile App development. For this, three issues to be determined are the representation of each candidate solution, the objective function, and the search strategy, which we define in the following before describing our proposed method step-by-step.
\end{comment}
\subsection{Representation of the candidate solutions}
\label{sec:rep}
Our proposed algorithm aims to find all hyperparameters in a MLNN. To the best of our knowledge, in none of the previous research studies, this set of hyperparameters has been optimised simultaneously. Our proposed neuro-evolution model can optimise the parameters and the number of layers and neurons in each layer. As a result, we propose a two-segment representation $N_{D}+N_{L}$. $N_{D}$ decision variables are responsible for finding the hyperparameters (except the number of neurons in each layer). In contrast, the second part is assigned to the number of neurons in each layer. It is crucial to note that the initial segment in our proposed representation has a constant length, while the subsequent segment exhibits a variable length.

The first segment of our proposed representation is allocated explicitly to all hyperparameters except the number of neurons, which is a string with a length of 9. The hyperparameters assigned to each decision variable, their types and values are given in Table~\ref{hyer_parameter}. We can see that all hyperparameters in the first segment are real values except the solver. A solver, here, refers to an optimisation algorithm used to train an MLNN, and it is responsible for finding the optimal set of weights and biases. However, there are different solvers in the literature. As a result, the proposed representation also fulfils the task of finding the proper solver among distinct solvers (here are 10). An integer value is assigned to this decision variable, ranging from 1 to 10, with each number denoting a specific solver. The solvers include adaptive moment estimation (Adam)~\cite{ADAM-ADAMMAX}, Adam with adaptive learning rate method (Adadelta)~\cite{Adadelta}, Adam with decoupled weight decay regularisation (AdamW)~\cite{ADAMW}, infinity norm-based ADAM (Adamax)~\cite{ADAM-ADAMMAX}, accelerated stochastic gradient descent (ASGD)~\cite{ASGD}, Nesterov momentum into Adam (NAdam)~\cite{NAdam},  rectified Adam (RAdam)~\cite{RAdam}, root mean square propagation (RMSprop)~\cite{RMSprop}, resilient backpropagation algorithm (Rprop)~\cite{RPROP} and stochastic gradient descent method (SGD)~\cite{SGD}, and each corresponds to a unique number from 1 to 10. Every solver within the experiment requires specific hyperparameters while disregarding others. For instance, the Adam algorithm requires the learning rate, $\beta_1$, $\beta_2$, and weight decay. Consequently, we employed a technique referred to as "\textbf{Selective Exclusion}" to calculate the objective function. Selective Exclusion entails the deliberate omission or exclusion of certain elements, a process contingent upon the chosen solver.

\begin{table}[]
	\centering
	\small
	\caption{Hyperparameters of the first segment of the proposed representation.}
	\begin{tabular}{l|c|l}
		\hline
		Hyperparameter        & type    &   values \\ \hline
		Learning rate   (lr)   & Real    & {[}0,1{]}         \\
		Weight decay           & Real    & {[}0,0.2{]}       \\
		$\rho$    & Real    & {[}0,1{]}         \\
		$\beta$1 & Real    & {[}0.8,1{]}       \\
		$\beta$2 & Real    & {[}0.8,1{]}       \\
		$\lambda$ & Real    & {[}0,1{]}         \\
		momentum               & Real    & {[}0,1{]}         \\
		Solver (optimiser)     & Integer & \{1,2,…,10\}  \\ \hline   
	\end{tabular}
	\label{hyer_parameter}
\end{table}

The subsequent segment of the proposed representation relates to allocating neurons in each hidden layer. It is important to note that the appropriate number of layers is controlled in the main algorithm. As a result, the size of the second segment varies according to the number of layers. For instance, if the number of layers is 4, the second segment will have a size of 4, with each decision variable in this segment corresponding to a distinct layer.

Therefore, we are tasked with determining appropriate values for 9 hyperparameters and the number of neurons in the hidden layers, encompassing a broad spectrum of possibilities within the search space. For example, consider the real-valued hyperparameters, which are discretised at intervals of \(0.01\). Additionally, assume that we have three layers and the maximum number of neurons in each layer is 50. This discretisation implies that, for the learning rate alone, there exist \(100\) potential values, though, in practice, the actual number of feasible values is greater. Consequently, the total number of potential combinations for the search space in this specific example is calculated as \(100^4 \times 20^3 \times 10 \times 50^3\). This exponential increase in possibilities underscores the complexity of identifying the optimal set, rendering the optimisation process notably challenging.

%\begin{algorithm2e}[tb]
%	\scriptsize
%	\SetAlgoLined
%	\SetKwInOut{Input}{Input}\SetKwInOut{Output}{Output}
%	\Input{ $D$: dimensionality of problem; $NFE_{\max}$: maximum number of function evaluations; $NP_{\max}$: maximum population size; $NP_{\min}$: minimum population size; $H$: Memory size}
%	\Output{ $x^*$: the best individual }
%	\BlankLine
%    Split the dataset into two parts: train ($TrainSet$) and test ($TestSet$) sets \;
%	\For{$NoHiddenLayer\leftarrow 1$ \KwTo $MaxHiddenLayer$} {
	%	Randomly generate initial population $Pop$\ based on the proposed representation and $NoHiddenLayer$ \;	
	%	Evaluate objective function value of each candidate solutions \;
	%	$model_{selected} \leftarrow$ Apply optimisation algorithm to find the best model \;
	%	\If{$f(model_{selected}) < f(model_{best}) $} {
		%	  $model_{best} \leftarrow  model_{selected}$
		%	}
	%	|}

%	\caption{The top-level view of the proposed Neuroevolution algorithm in the form of Pseudo-code.}
%	\label{Alg1:proposed}
%\end{algorithm2e}

\subsection{Concealing Strategy}
In real-world mobile Apps, if a user does not grant permissions for certain features, those features may be missing from the feature set. In other words, in a typical situation, it is likely that there will be missing values in the data. However, while there are several approaches for handling missing values, this paper does not focus on identifying the best one. Instead, our proposed method integrates a concealing (masking) strategy into the MLNN to handle missing values. The concealing strategy employs binary masks to address missing values within the input data. These binary masks serve as indicators, where 1 represents a known or observed feature, while 0 indicates a missing or unobserved feature.

Mathematically, we define $x \in \mathbb{R}^{p}$ as a vector representing an input sample comprising $p$ features. We assume that the features in $x$ can be divided into two distinct sets: $q$ observed features denoted as $x^{o} \in \mathbb{R}^{q}$, and $r$ missing features denoted as $x^{m}$, where the sum of observed and missing features equals the total number of features, $p$ ($q+r=p$). In our concealing strategy, the activation function of the $k$-th ($k=1,2,...,s$) neuron in the first hidden layer of an MLNN is defined as
\begin{equation}
	a_k^{(1)} = f\left(\sum_{i=1}^{q} w_{ik}^{(1)} x_i^{o} + \sum_{j=1}^{r} w_{jk}^{(1)} x_j^{m} + b_k^{(1)}\right)
\end{equation}

where$a_k^{(1)}$ denotes the activation of the $k$-th neuron in the first hidden layer,
$f$ represents the chosen activation function (e.g., sigmoid, ReLU, etc.),
$\sum_{i=1}^{q} w_{ik}^{(1)} x_i^{o}$ is the weighted sum of observed features from the input, where $x_i^{o}$ is the $i$-th observed feature, and $w_{ik}^{(1)}$ is the weight associated with the connection between the $i$-th observed feature and the $k$-th neuron in the first hidden layer.
$\sum_{j=1}^{r} w_{jk}^{(1)} x_j^{m}$ is the weighted sum of missing features from the input, where $x_j^{m}$ is the j-th missing feature, and $w_{jk}^{(1)}$ is the weight associated with the connection between the $j$-th missing feature and the $k$-th neuron in the first hidden layer. Also, $b_k^{(1)}$ represents the bias term for the $k$-th neuron in the first hidden layer.

In our proposed approach, the missing values in the data set are represented by a binary mask matrix $M \in \{0, 1\}^{n \times p}$, where $n$ is the number of samples.  For each sample \(i\) and feature \(j\), the mask value \(M_{ij}\) is set as follows:
\begin{equation}
	M_{ij} = 
	\begin{cases} 
		1 & \text{if feature } j \text{ in sample } i \text{ is observed}, \\
		0 & \text{if feature } j \text{ in sample } i \text{ is missing}.
	\end{cases}
\end{equation}

Hence, for each sample $i$, $M_i = (M_{i1}, M_{i2}, ..., M_{ip})$ represents the binary mask vector indicating missing (0) and observed (1) features.

%The binary mask matrix $M$ is defined as follows: if the $j$-th feature of the $i$-th sample is missing, $M_{ij}$ takes the value 0, else it takes the value 1. Hence, for each sample $i$, $M_i = (M_{i1}, M_{i2}, ..., M_{ip})$ represents the binary mask vector indicating missing (0) and observed (1) features.

% To apply the masking approach, the original feature vectors are multiplied element-wise with the binary mask matrix: $x_i^{m} = x_i \odot M_i$, where $x_i^{m} \in \mathbb{R}^p$ represents the masked feature vector for the $i$-th sample. 

To incorporate the concealing strategy into the MLNN, we compute the masked input \(x_{i}^{m}\) for each sample \(i\) by performing an element-wise multiplication between the original input vector \(x_{i}\) and its corresponding binary mask vector \(M_{i}\), yielding:
\begin{equation}
	x_{i}^{m} = x_{i} \odot M_{i},
\end{equation}
where \(\odot\) denotes the element-wise multiplication operation, and \(x_{i}^{m} \in \mathbb{R}^{p}\) represents the masked input vector with missing values concealed. By performing the element-wise multiplication with the binary mask, the missing values are effectively masked or set to zero, while the observed values remain unchanged. This allows subsequent analysis, such as the calculation of activation function, since the activation function cannot be calculated in the state where there are missing values.

We have used the concealing strategy here since 1) it preserves the data structure by keeping track of missing values while retaining the remaining observed values, 2) it retains the sample size since by masking missing values, the original sample size is maintained, ensuring that all available data points are utilised in the analysis, and 3) it helps to prevent biased estimates or erroneous conclusions that could arise from ignoring missing values, leading to more accurate predictions. It is worth mentioning that the concealing strategy is similar to imputing missing values by 0. Although in concealing method, the network knows what values have been missed, which may be useful for future applications.

% It is worth mentioning that the concealing strategy differs from imputing missing values by 0. In particular, when imputing missing values with 0 (or another value), all weights, including those corresponding to imputed values, are updated during training, while in masking to handle missing values, weights associated with the masked (missing) values are not updated during training, as these values are effectively excluded from the computation of gradients and weight updates. As a result, it inherently reduces bias since it excludes missing values from the computation. This ensures that the network does not infer relationships from non-existent data points, promoting a more accurate representation of the underlying patterns. Imputing with 0 can introduce bias, as the network may inadvertently learn associations between the value 0 and specific patterns that might not exist in the data. This could lead to skewed conclusions and affect the model's overall performance, mainly when missing values carry essential information.

\subsection{Objective Function}
\label{sec:obj}
The objective function in this study indicates the quality of a specific architecture designed for energy consumption in our MLNN model. The objective function here is based on the classification accuracy, which is calculated based on the neural network's predictions, considering the concealing strategy for handling missing values. The classification error is computed as:
\begin{equation}
	\text{Classification error}=\frac{100}{P} \times \sum_{p=1}^{P} \xi(x_{p})
	\label{Eq_Obj}
\end{equation}
with
\begin{equation}
	\xi(x_{p})=
	\begin{cases}
		1 &  \text{if} \hspace{0.1cm} o_{p}\neq d_{p}\\
		0& \hspace{-0.04cm} \text{otherwise}
	\end{cases}\hspace{2cm}\\
	\label{eq7}
\end{equation}
where $P$ is the number of samples, and for each input vector in training data, the corresponding desired output is $d_{p}$, and $o_{p}$ is the predicted output by the neural network. To ensure reliable and robust results, the objective function is evaluated using $k$-fold cross-validation.

\section{Results and Discussions }
\label{sec:results}
In this section, we conducted a comprehensive set of experiments to evaluate the effectiveness of our proposed approach. To achieve this, our evaluation focuses on two primary metrics: accuracy and F-measure. Also, to ensure robustness, the whole process is repeated 10 times and statistics such as mean and standard deviation are reported. A detailed description of the data set we used can be found in Section~\ref{sec:dataset}. For our evaluation, we have defined two scenarios: 1) evaluation in the case where there is no missing value, and 2) assessment in the case where there are missing values in the data set.

\subsection{Data set}
\label{sec:dataset}

In this study, we utilised the GreenHub data set~\cite{greenhub_dataset} as the foundation of our research. This comprehensive data set comprises over 23 million instances, spanning 900 brands and 5,000 models and covering 160 countries. It encompasses three distinct types of information: the sample data set, device data set, and App processes data set. The sample data set presents various details on different smartphone settings, while the device data set includes features related to smartphone brands and specifications. On the other hand, the App processes data set contains information about the installed applications on each smartphone.

Our research focused solely on the sample data set, deliberately disregarding the App processes and device data sets. This decision proposes a predictive algorithm that remains independent of the installed applications. Incorporating App processes would introduce challenges due to the regional variations in commonly used applications. For instance, while "WeChat" may be prevalent in China, "WhatsApp" may hold that position in Europe and in the US. Consequently, including such features would lead to sparsity in the data set, making the predictive process complex and less reliable. By excluding this data, we can ensure the robustness and geographical independence of the results. Similarly, the device data set also presents similar challenges, which led us to focus solely on Android settings. This approach ensures that our results are independent of varying device specifications across regions.

The sample data set contains seven distinct features, from which we selected 32 relevant features for our study. These features encompass various aspects such as battery details (charger status, health, voltage, and temperature), CPU states (usage, uptime, and sleep time), network details (network type, mobile network type, mobile data status, mobile data activity, roaming enabled, wifi status, wifi signal strength, and wifi link speed), samples (battery state, battery level, memory free, memory user, network status, screen brightness, and screen on), settings (Bluetooth enabled, location enabled, power saver enabled, flashlight enabled, NFC enabled, and developer mode), and storage details (free and total memory, memory active, and memory inactive). Additional information can be referenced from \cite{greenhub_dataset} for further specifics on these features.

\subsection{First Scenario}
In this section, we have evaluated our proposed approach in the case where there are no missing values in our data set. To this end, the proposed approach is integrated with 13 PBMHs. While the representation and objective functions remained consistent across all algorithms, each algorithm employed a distinct search strategy. Also, the population size for all algorithms is set at 10. Since the objective function in the proposed algorithm is computationally expensive, we used a limited number of fitness evaluations for the proposed algorithm. In other words, the number of function evaluations for the optimisation process for each layer is selected as 30, and the maximum number of layers is 8. The remaining parameters, as listed in Table 1 in the \textbf{Supplementary File}, are set to their default values.

Table~\ref{tab:results_without_missing} compares the performance of various algorithms using the accuracy and F-measure metrics. Among the algorithms evaluated, PSO-AMLNN, SAPE-DE-AMLNN, and DE-AMLNN emerge as the top three performers based on accuracy and F-measure metrics. PSO-AMLNN achieves the highest mean accuracy (87.63) and F-measure (87.96) scores, demonstrating its effectiveness in accurately classifying the test data and achieving a balance between precision and recall. SAPE-DE follows closely behind, with mean accuracy (87.38) and F-measure (87.26) scores, indicating its strong predictive capabilities. 
%DE also shows notable performance, with mean accuracy (84.75) and F-measure (84.43) scores, suggesting its effectiveness in classification tasks.

On the other hand, GA-AMLNN, JADE-AMLNN, and SADE-AMLNN exhibit relatively weaker performance among the evaluated algorithms. GA-AMLNN presents the lowest mean accuracy (72.75) and F-measure (72.60) scores. JADE-AMLNN also can not perform well compared to others, and it is the second-worst algorithm. SADE-AMLNN performs relatively better than GA-AMLNN and JADE-AMLNN but still falls behind the top-performing algorithms, with mean accuracy (80.25) and F-measure (80.38) scores.

\begin{table}[!htb]
	\scriptsize\setlength{\extrarowheight}{2pt}
	\centering
	\caption{Experimental results for the first scenario on 10 independent iterations.}
	\label{tab:results_without_missing}	
	\begin{tabular}{l|cc|cc}
		\hline
		\multirow{2}{*}{Algorithms} & \multicolumn{2}{c}{Accuracy} & \multicolumn{2}{c}{F-measure} \\ \cline{2-5} 
		& Mean          & Std.         & Mean           & Std.         \\ \hline
		GA-AMLNN                          & 72.75         & 1.50         & 72.60          & 1.42         \\
		DE-AMLNN                          & 84.75         & 2.08         & 84.43          & 2.15         \\
		MA-AMLNN                          & 82.00         & 1.83         & 81.36          & 1.92         \\
		PSO-AMLNN                         & 87.63         & 2.04         & 87.96          & 2.15         \\
		CMA-ES-AMLNN                      & 81.63         & 1.59         & 81.35          & 1.63         \\	
		HPSO-AMLNN                   & 84.06         & 2.12         & 84.36          & 2.11         \\	
		CPSO-AMLNN                        & 85.69         & 2.41         & 85.65          & 2.30         \\	
		CLPSO-AMLNN                       & 84.56         & 2.16         & 84.65          & 2.11         \\		
		SAPE-DE-AMLNN                     & 87.38         & 1.80         & 87.26          & 1.42         \\
		JADE-AMLNN                        & 76.94         & 1.88         & 76.95          & 1.47         \\
		SHADE-AMLNN                       & 84.38         & 2.35         & 84.91          & 2.93         \\
		LSHADE-AMLNN                     & 82.38         & 2.18         & 82.81          & 2.16         \\		
		PPSO-AMLNN                        & 84.81         & 1.24         & 84.99          & 1.30         \\
		\hline
	\end{tabular}
\end{table}

To have a visual understanding, we show the generated neural network architectures found by using each algorithm in Table~\ref{tab:architecture} in terms of the three most important parameters that are common in all solvers, which are the configuration of neurons in the hidden layers of the neural network, along with the specified learning rate and solver used during training.

From the table, the neural network architectures vary across the algorithms, demonstrating the diverse approaches employed by each algorithm to solve the problem. The number of neurons in the hidden layers varies significantly, ranging from 173 to 400. This disparity suggests that different algorithms adopt different levels of complexity in their neural network structures to address the task. Furthermore, each algorithm's learning rate, which determines the step size in weight adjustments during training, is also distinct. The specified learning rates range from 0.01 to 0.17, indicating varying sensitivity to weight updates during training. A lower learning rate implies more cautious adjustments, while a higher learning rate allows for more significant changes. In addition, the solver, Rprop, is found consistently across all algorithms in this study.  

Comparing Table~\ref{tab:results_without_missing} (numerical results) and Table~\ref{tab:architecture} (generated architectures), we can identify potential relationships between algorithm performance and neural network architectures. For instance, we can observe that PSO-AMLNN and SAPE-DE-AMLNN, which achieved high accuracy and F-measure scores in Table~\ref{tab:results_without_missing}, have relatively complex structures with multiple hidden layers and a larger number of neurons. This complexity may enable them to capture complicated relationships within the data, resulting in improved performance. Conversely, GA-AMLNN, with a simpler architecture, has lower performance.

Additionally, the learning rate can also influence algorithm performance and convergence. Algorithms with lower learning rates, such as SAPE-DE-AMLNN and PSO-AMLNN, might exhibit a more cautious learning behaviour, gradually adjusting their weights to achieve optimal performance. On the other hand, algorithms with higher learning rates, like GA-AMLNN, may experience more abrupt weight updates, potentially affecting convergence and final performance.

\begin{table}[!htb]
	\scriptsize\setlength{\extrarowheight}{2pt}
	\centering
	\caption{The architecture found by each algorithm in the first scenario.}
	\label{tab:architecture}
	\addtolength{\tabcolsep}{-3.8pt}	
	\begin{tabular}{l|l|c|c}
		Algorithm    & Structure                    & Leaning rate & Solver \\ \hline
		GA-AMLNN      & {[}223{]}                       & 0.17         & Rprop  \\
		DE-AMLNN       & {[}302,11{]}                    & 0.05         & Rprop  \\
		MA-AMLNN       & {[}366,112{]}                   & 0.04         & Rprop  \\
		PSO-AMLNN      & {[}173,262,294,12{]}            & 0.01         & Rprop  \\
		CMA-ES-AMLNN   & {[}343,18,367{]}                & 0.01         & Rprop  \\
		HPSO-AMLNN    & {[}217{]}                       & 0.02         & Rprop  \\
		CPSO-AMLNN    & {[}202,226{]}                   & 0.01         & Rprop  \\
		CLPSO-AMLNN   & {[}220,21,356{]}                & 0.01         & Rprop  \\
		SAPE-DE-AMLNN & {[}278,39,186,295,324{]}        & 0.01         & Rprop  \\
		JADE-AMLNN    & {[}272{]}                       & 0.1          & Rprop  \\
		SHADE-AMLNN   & {[}400{]}                       & 0.05         & Rprop  \\
		LSHADE-AMLNN  & {[}294,392,221,   21,243,268{]} & 0.02         & Rprop  \\
		PPSO-AMLNN    & {[}183,358,103,75,104,323{]}    & 0.01         & Rprop \\ \hline
	\end{tabular}
\end{table}

\subsection{Second Scenario}

This section provides our proposed algorithm's results when there are missing values in the data set. In the initial test, we introduced a scenario where 5\%  of the total elements are deliberately excluded. To illustrate, considering a data set with 8,000 samples and 23 available features, this implies that 920 elements out of a total of 18,400 elements are randomly omitted. For the experiments, we used the same parameters with the first scenario . 

Table~\ref{tab:5percent} gives the results of our proposed algorithm on the data set with 5\% missing values and 10 independent iterations. By analysing Table~\ref{tab:5percent}, we observe variations in the performance of algorithms when dealing with missing values. For instance, GA-AMLNN shows a decrease in both accuracy and F-measure compared to its performance in the first scenario without any missing values. This suggests that GA-AMLNN is more sensitive to the presence of missing values. In contrast, some algorithms demonstrate a more robust performance despite the missing values. CLPSO-AMLNN, SADE-AMLNN, SAPE-DE-AMLNN, and JADE-AMLNN show consistent mean accuracy and F-measure scores, indicating their ability to handle missing data and provide reliable predictions. 

Comparing the results between the two tables highlights the importance of algorithm selection in the presence of missing values. Algorithms like CLPSO-AMLNN, SADE-AMLNN, SAPE-DE-AMLNN, and JADE-AMLNN maintain their effectiveness even when confronted with missing values. Conversely, GA-AMLNN and PSO-AMLNN demonstrate notable decreases in performance, indicating their vulnerability to the presence of missing data.

Table~\ref{tab:5architecture} presents the architectures found by each algorithm in the original data set with 5\% missing values. Analysing Table~\ref{tab:5architecture} provides insights into how the presence of missing values affects the generated architectures. One interesting pattern is that the average number of neurons in the architectures obtained from Table~\ref{tab:5architecture} is slightly lower when compared to Table~\ref{tab:architecture}, indicating potential adaptations to account for the presence of missing data. For example, GA-AMLNN exhibits a structure of {[}128{]} in Table~\ref{tab:5architecture}, whereas it had {[}223{]} neurons in Table~\ref{tab:architecture}. In addition, all algorithms, except HPSO-AMLNN and PPSO-AMLNN, suggest Rprop for their solvers, while these two proposes AdamW as their learning algorithms.

\begin{table}[!htb]
	\scriptsize\setlength{\extrarowheight}{2pt}
	\centering
	\caption{Experimental results in the second scenario with 5\%  missing values on 10 independent iterations.}
	\label{tab:5percent}
	\begin{tabular}{l|cc|cc}
		\hline
		\multirow{2}{*}{Algorithms} & \multicolumn{2}{c}{Accuracy} & \multicolumn{2}{c}{F-measure} \\ \cline{2-5} 
		& Mean          & Std.         & Mean           & Std.         \\ \hline
		GA-AMLNN                          & 64.75         & 1.08         & 64.81          & 1.05         \\
		DE-AMLNN                          & 78.25         & 2.44         & 78.27          & 2.41         \\
		MA-AMLNN                          & 77.19         & 2.87         & 77.17          & 2.92         \\
		PSO-AMLNN                         & 64.19         & 1.63         & 77.17          & 2.92         \\
		CMA-ES-AMLNN                      & 83.25         & 2.31         & 83.30          & 2.30         \\	
		HPSO-AMLNN                  & 72.50         & 1.73         & 72.44          & 1.77         \\
		CPSO-AMLNN                        & 72.75         & 1.13         & 72.79          & 1.10         \\	
		CLPSO-AMLNN                       & 85.06         & 2.03         & 85.05          & 2.03         \\	
		SAPE-DE-AMLNN                     & 83.75         & 2.85         & 83.74          & 2.87         \\
		JADE-AMLNN                        & 85.25         & 2.45         & 85.24          & 2.45         \\
		SHADE-AMLNN                       & 74.63         & 2.83         & 74.57          & 2.83         \\
		LSHADE-AMLNN                     & 77.44         & 2.47         & 77.37          & 2.62         \\ 
		PPSO-AMLNN                        & 73.13         & 1.19         & 73.08          & 1.22         \\ \hline

	\end{tabular}
\end{table}

\begin{table}[!htb]
	\scriptsize\setlength{\extrarowheight}{2pt}
	\centering
	\caption{The architecture found by each algorithm in the second scenario with 5\% missing values.}
	\label{tab:5architecture}
	\begin{tabular}{l|l|c|c}
		Algorithm    & Structure     & Leaning rate & Solver \\ \hline
		GA-AMLNN      & {[}128{]}        & 0.18         & Rprop  \\
		DE-AMLNN      & {[}287{]}        & 0.09         & Rprop  \\
		MA-AMLNN     & {[}187{]}        & 0.05         & Rprop  \\
		PSO-AMLNN     & {[}98{]}         & 0.14         & Rprop  \\
		CMA-ES-AMLNN  & {[}282,68{]}     & 0.01         & Rprop  \\
		HPSO-AMLNN    & {[}214{]}        & 002          & AdamW  \\
		CPSO-AMLNN    & {[}219{]}        & 0.13         & Rprop  \\
		CLPSO-AMLNN   & {[}320{]}        & 0.01         & Rprop  \\
		SAPE-DE-AMLNN & {[}260{]}        & 0.01         & Rprop  \\
		JADE-AMLNN    & {[}356,84,133{]} & 0.01         & Rprop  \\
		SHADE-AMLNN   & {[}282{]}        & 0.09         & Rprop  \\
		LSHADE-AMLNN  & {[}397{]}        & 0.1          & Rprop  \\
		PPSO-AMLNN    & {[}174{]}        & 0.02         & AdamW \\ \hline 
	\end{tabular}
\end{table}

In the next experiment, we increased the missing values to 20\%. Table~\ref{tab:20percent} provides the results of experiments in the data set with 20\% missing values. Analysing Table~\ref{tab:20percent} allows for a discussion on the impact of increased missing value in the data set and a comparison with the previous two tables (Table~\ref{tab:5percent}: 5\% missing values, and Table~\ref{tab:results_without_missing}: original data set without missing values). With the inclusion of 20\% missing values, we observe noticeable changes in algorithm performance. Some algorithms, such as GA-AMLNN, MA-AMLNN, PSO-AMLNN, SAPE-DE-AMLNN, and SHADE-AMLNN, experience decreased mean accuracy and F-measure scores compared to their performance in Table~\ref{tab:5percent} (5\% missing values). This suggests that these algorithms are more sensitive to increased missing values, leading to reduced prediction accuracy. In contrast, several algorithms, including DE-AMLNN, CMA-ES-AMLNN, CPSO-AMLNN, HPSO-AMLNN, SADE-AMLNN, JADE-AMLNN, and LSHADE-AMLNN, maintain relatively stable mean accuracy and F-measure scores in the presence of 20\% missing values. This indicates their robustness to handle increased noise and their ability to generate reliable predictions. In addition, comparing Table~\ref{tab:20percent} with Table~\ref{tab:results_without_missing} (original data set without missing values) reveals the overall impact of missing values on algorithm performance. Most algorithms in Table~\ref{tab:20percent} exhibit lower mean accuracy and F-measure scores compared to their counterparts in Table~\ref{tab:5percent}. %This demonstrates the challenges posed by missing values, as they can degrade the performance of the algorithms in predicting energy consumption accurately.

\begin{table}[!htb]
	\scriptsize\setlength{\extrarowheight}{2pt}
	\centering
	\caption{Experimental Results in the second scenario with 20\%  missing values.}
	\label{tab:20percent}
	\begin{tabular}{l|cc|cc}
		\hline
		\multirow{2}{*}{Algorithms} & \multicolumn{2}{c}{Accuracy} & \multicolumn{2}{c}{F-measure} \\ \cline{2-5} 
		& Mean          & Std.         & Mean           & Std.         \\ \hline
		GA-AMLNN                           & 71.69         & 2.98         & 71.67          & 2.97         \\
		DE-AMLNN                           & 78.94         & 2.77         & 78.86          & 2.87         \\
		MA-AMLNN                           & 65.63         & 2.15         & 65.62          & 2.19         \\
		PSO-AMLNN                          & 70.19         & 2.52         & 69.99          & 2.51         \\		
		CMA-ES-AMLNN                       & 84.50         & 2.38         & 84.54          & 2.34         \\
		HPSO-AMLNN                    & 77.19         & 2.18         & 77.16          & 2.21         \\
		CPSO-AMLNN                         & 79.13         & 2.28         & 79.13          & 2.33         \\
		CLPSO-AMLNN                        & 77.75         & 2.93         & 77.73          & 2.92         \\
		SAPE-DE-AMLNN                      & 68.88         & 2.12         & 68.83          & 2.20         \\
		JADE-AMLNN                         & 80.88         & 2.52         & 80.77          & 2.65         \\
		SHADE-AMLNN                        & 68.38         & 1.81         & 68.49          & 1.73         \\
		LSHADE-AMLNN                      & 76.13         & 2.39         & 76.09          & 2.37         \\ 		
		PPSO-AMLNN                         & 79.94         & 2.71         & 79.83          & 2.91         \\ \hline
		
	\end{tabular}
\end{table}

Table~\ref{tab:20architecture} shows the architectures found by each algorithm in the second scenario with 20\% missing values. The structures in Table~\ref{tab:20architecture} exhibit variations compared to Table~\ref{tab:5architecture}, indicating algorithmic adaptations to handle the higher missing values. Certain algorithms, such as GA-AMLNN, DE-AMLNN, and MA-AMLNN, demonstrate changes in their neural network structures to adjust for the increased missing values. Such alterations also exit for the learning rate and solver.
%, which reflects the algorithms' attempts to capture the underlying patterns in the presence of higher missing value levels.  

\begin{table}[!htb]
	\scriptsize\setlength{\extrarowheight}{2pt}
	\centering
	\caption{The architecture found by each algorithm in the second scenario with 20\% missing values .}
	\label{tab:20architecture}
	\addtolength{\tabcolsep}{-4.2pt}
	\begin{tabular}{l|l|c|c}
		\hline
		Algorithm    & Structure                     & Leaning rate & Solver \\ \hline
		GA-AMLNN      & {[}110, 140{]}                   & 0.07         & Rprop      \\
		DE-AMLNN      & {[}149{]}                        & 0.01         & Rprop      \\
		MA-AMLNN      & {[}81{]}                         & 0.11          & Rprop      \\
		PSO-AMLNN     & {[}187{]}                        & 0.02         & AdamW      \\
		CMA-ES-AMLNN  & {[}342,21,137{]}                 & 0.01         & Rprop      \\
		HPSO-AMLNN    & {[}350{]}                        & 0.11         & Rprop      \\
		CPSO-AMLNN    & {[}126{]}                        & 0.02         & Rprop      \\
		CLPSO-AMLNN   & {[}148,14{]}                     & 0.07         & Rprop      \\
		SAPE-DE-AMLNN & {[}55{]}                         & 0.01         & Rprop      \\
		JADE-AMLNN    & {[}392,63,6,291,212,299,160{]}   & 0.05         & Rprop      \\
		SHADE-AMLNN   & {[}193{]}                        & 0.20         & Rprop      \\
		LSHADE-AMLNN  & {[}199{]}                        & 0.08         & Rprop      \\
		PPSO-AMLNN    & {[}313,89,201,344,376,365,344{]} & 0.01         & AdamW   \\ \hline  
	\end{tabular}
\end{table}

In the next experiment, we conducted experiments under severe noise conditions, with 40\% of the values being missed. The results in Table~\ref{tab:40percent} indicate a considerable decrease in algorithm performance compared to the previous tables. The mean accuracy and F-measure scores for most algorithms, such as CPSO-AMLNN, SAPE-DE-AMLNN, and LSHADE-AMLNN,  are noticeably lower, signifying the challenges introduced by severe noise in the data set. On the other hand, some algorithms, such as CMA-ES-AMLNN and CLPSO-AMLNN, demonstrate relatively stable mean accuracy and F-measure scores even in the presence of severe noise. This resilience indicates the algorithms' effectiveness in making reliable predictions.

\begin{table}[!htb]
	\scriptsize\setlength{\extrarowheight}{2pt}
	\centering
	\caption{Experimental Results in the second scenario with 40\%  missing values.}
	\label{tab:40percent}
	\begin{tabular}{l|cc|cc}
		\hline
		\multirow{2}{*}{Algorithms} & \multicolumn{2}{c}{Accuracy} & \multicolumn{2}{c}{F-measure} \\ \cline{2-5} 
		& Mean          & Std.         & Mean           & Std.         \\ \hline
		GA-AMLNN                          & 78.38         & 2.92         & 78.38          & 2.92         \\
		DE-AMLNN                          & 65.25         & 1.32         & 65.39          & 1.47         \\
		MA-AMLNN                          & 74.81         & 2.19         & 74.76          & 2.17         \\
		PSO-AMLNN                         & 71.50         & 2.92         & 71.39          & 2.01         \\		
		CMA-ES-AMLNN                      & 83.50         & 2.21         & 83.50          & 2.22         \\
		HPSO-AMLNN                   & 62.94         & 2.38         & 62.90          & 2.43         \\
		CPSO-AMLNN                        & 52.56         & 1.09         & 52.44          & 1.24         \\
		CLPSO-AMLNN                       & 79.81         & 2.92         & 79.83          & 2.94         \\
		SAPE-DE-AMLNN                     & 51.31         & 1.43         & 51.19          & 1.31         \\
		JADE-AMLNN                        & 67.94         & 2.74         & 67.92          & 2.78         \\
		SHADE-AMLNN                       & 62.69         & 2.00         & 62.76          & 2.08         \\
		LSHADE-AMLNN                     & 48.75         & 1.25         & 48.70          & 1.20         \\ 
		PPSO-AMLNN                        & 70.81         & 2.80         & 70.80          & 2.85         \\ \hline	
	\end{tabular}
\end{table}
Table~\ref{tab:40architecture} indicates the architectural configurations obtained by each algorithm when applied to the original data set with a 40\% rate of missing values. The quantitative analysis of the architectural configurations, learning rates, and solver choices provides insights into the strategies employed by the algorithms to handle severe noise. The observed variations reflect the algorithms' adaptive behaviours and attempts to optimise performance under challenging data conditions.
\begin{table}[!htb]
	\scriptsize\setlength{\extrarowheight}{2pt}
	\centering
	\caption{The architecture found by each algorithm in the second scenario with 40\% missing values.}
	\label{tab:40architecture}
	\addtolength{\tabcolsep}{-3.8pt}
	\begin{tabular}{l|l|c|c} 
		\hline
		Algorithm    & Architecture                      & Leaning rate & Solver \\ \hline
		GA-AMLNN      & {[}318,55{]}                      & 0.01         & AdamW  \\
		DE-AMLNN      & {[}124{]}                         & 0.14         & Rprop  \\
		MA-AMLNN      & {[}94{]}                          & 0.01         & Rprop  \\
		PSO-AMLNN     & {[}400,64,400,128,224,110{]}      & 0.01         & AdamW  \\
		CMA-ES-AMLNN  & {[}247{]}                         & 0.12         & Rprop  \\
		HPSO-AMLNN    & {[}133{]}                         & 0.14         & Rprop  \\
		CPSO-AMLNN    & {[}11{]}                          & 0.09         & Rprop  \\
		CLPSO-AMLNN   & {[}224,287,307,391,124,294,292{]} & 0.01         & Rprop  \\
		SAPE-DE-AMLNN & {]}10{]}                          & 0.11         & Rprop  \\
		JADE-AMLNN    & {[}173,250,95,185{]}              & 0.01         & Rprop  \\
		SHADE-AMLNN   & {[}105{]}                         & 0.17         & Rprop  \\
		LSHADE-AMLNN  & {[}4{]}                           & 0.05         & Rprop  \\
		PPSO-AMLNN    & {[}113{]}                         & 0.07         & Rprop \\ \hline 
	\end{tabular}
\end{table}

\begin{figure}[!htb]
	\centering
	\includegraphics[width=0.9\columnwidth]{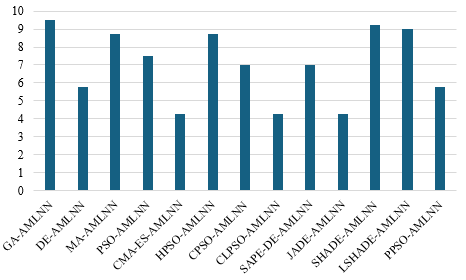}
	\caption{Results of Friedman test.} 
	\label{tab:friedman_acc}
\end{figure}
\begin{comment}

\begin{table}[!htb]
	\scriptsize\setlength{\extrarowheight}{2pt}
	\centering
	\caption{Results of Friedman test.}
	\label{tab:friedman_acc}
	\begin{tabular}{l|cc}
		Algorithms   & Average rank   & Overall rank   \\ \hline
		GA-AMLNN      & 9.5            & 13             \\
		DE-AMLNN      & 5.75           & 4.5            \\
		MA-AMLNN      & 8.75           & 9.5            \\
		PSO-AMLNN     & 7.50           & 8              \\
		CMA-ES-AMLNN  & 4.25           & 2              \\
		HPSO-AMLNN    & 8.75           & 9.5            \\
		CPSO-AMLNN    & 7.00           & 6.5            \\
		CLPSO-AMLNN   & 4.25           & 2              \\
		SAPE-DE-AMLNN & 7.00           & 6.5            \\
		JADE-AMLNN    & 4.25           & 2              \\
		SHADE-AMLNN   & 9.25           & 12             \\
		LSHADE-AMLNN  & 9.00           & 11             \\
		PPSO-AMLNN    & 5.75           & 4.5            \\ \hline
		Chi-square   & \multicolumn{2}{c}{61.48}       \\
		p-value      & \multicolumn{2}{c}{1.20869E-08} \\ \hline
	\end{tabular}
\end{table}
\end{comment}

\begin{sidewaystable}[!phtb]
	\scriptsize\setlength{\extrarowheight}{1.8pt}
	\centering
	\setlength{\tabcolsep}{1.1pt}
	\caption{
		The results obtained from the Wilcoxon signed-rank test are based on the mean objective function value. Symbols $+$, $-$, and $=$ are used to indicate whether an algorithm is statistically equivalent to, statistically superior to, or statistically inferior to another algorithm, respectively. Additionally, the last column summarises the cumulative wins (w), ties (t), and losses (l) of each algorithm.}
	\label{tab:wil}
	\begin{tabular}{l|c|c|c|c|c|c|c|c|c|c|c|c|c|l}
		Algorithms   & GA-AMLNN & DE-AMLNN & MA-AMLNN & PSO-AMLNN & CMA-ES-AMLNN & HPSO-AMLNN & CPSO-AMLNN & CLPSO-AMLNN & SAPE-DE-AMLNN & JADE-AMLNN & SHADE-AMLNN & LSHADE-AMLNN & PPSO-AMLNN & w/t/l  \\ \hline
		GA-AMLNN      &\cellcolor{gray!25}  & -       & -       & -        & -           & -         & -         & -          & -            & -         & -          & =           & -         & 0/1/11 \\
		DE-AMLNN      & +       & \cellcolor{gray!25} & +       & +        & -           & +         & +         & -          & +            & -         & +          & +           & =         & 8/1/3  \\
		MA-AMLNN      & +       & -       & \cellcolor{gray!25}  & -        & -           & =         & -         & -          & -            & -         & +          & +           & -         & 3/1/8  \\
		PSO-AMLNN     & +       & -       & +       & \cellcolor{gray!25}  & =           & =         & -         & =          & -            & -         & +          & +           & -         & 4/3/5  \\
		CMA-ES-AMLNN  & +       & +       & +       & =        & \cellcolor{gray!25}  & =         & +         & =          & =            & =         & +          & +           & +         & 7/5/0  \\
		HPSO-AMLNN    & +       & -       & =       & =        & =           & \cellcolor{gray!25}      & -         & =          & -            & -         & +          & +           & -         & 3/4/5  \\
		CPSO-AMLNN    & +       & -       & +       & +        & -           & +         &\cellcolor{gray!25}       & =          & =            & =         & +          & +           & -         & 6/3/3  \\
		CLPSO-AMLNN   & +       & +       & +       & =        & =           & =         & =         &\cellcolor{gray!25}        & =            & =         & +          & +           & +         & 6/6/0  \\
		SAPE-DE-AMLNN & +       & -       & +       & +        & =           & +         & =         & =          &\cellcolor{gray!25}          & =         & +          & +           & -         & 6/4/2  \\
		JADE-AMLNN    & +       & +       & +       & +        & =           & +         & =         & =          & =            & \cellcolor{gray!25}      & +          & +           & +         & 8/4/0  \\
		SHADE-AMLNN   & +       & -       & -       & -        & -           & -         & -         & -          & -            & -         & \cellcolor{gray!25}       & -           & -         & 1/0/11 \\
		LSHADE-AMLNN  & =       & -       & -       & -        & -           & -         & -         & -          & -            & -         & +          &  \cellcolor{gray!25}       & -         & 1/0/10 \\
		PPSO-AMLNN    & +       & =       & +       & +        & -           & +         & +         & -          & +            & -         & +          & +           & \cellcolor{gray!25}      & 8/1/3 \\ \hline
	\end{tabular}
\end{sidewaystable}

\begin{figure}[!htb]
	\centering
	\includegraphics[width=0.9\columnwidth]{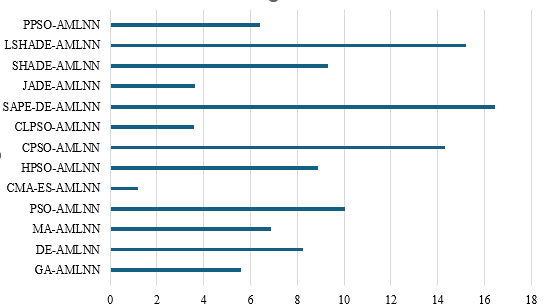}
	\caption{Stability results.} 
	\label{tab:std}
\end{figure}

\subsection{ Overall Analysis}
Statistical analysis is crucial in drawing meaningful conclusions from data and making informed decisions in PBMH algorithms. 
%In this case, the alternative hypothesis, denoted as $H_{1}$, suggests the presence of a meaningful variation among the algorithms in question. On the other hand, the null hypothesis, referred to as $H_{0}$, states that there is no statistically significant difference among the algorithms. $H_{0}$ is the initial assumption that is tested, and only if the $H_{0}$
%is shown to be false, the alternative hypothesis $H_{1}$ is accepted. Generally speaking, there are two types of statistical tests: multiple, for comparison of more than two samples, and pairwise tests, for comparison of two algorithms. To this end, the Wilcoxon signed-rank test, a pairwise test, and the Friedman test, a multiple test, are two important non-parametric statistical tests~\cite{tutorial_statistical}. 

The Friedman test is a non-parametric alternative to the one-way ANOVA (Analysis of Variance) and is used to compare multiple related samples simultaneously. 
%In the Friedman test, data is collected for multiple groups under different conditions. Each group is ranked independently across all states, and the ranks are used to calculate the test statistic. 
The test determines whether there are significant differences in the rankings among the groups. If the p-value obtained from the Friedman test is below a chosen significance level, $\alpha$, it indicates significant differences among the groups. The results of the Friedman test, as shown in Figure~\ref{tab:friedman_acc}, provide valuable insights into the relative performance of various PBMH algorithms. The primary objective of this study was to compare and rank these algorithms based on their accuracy results. According to the ranking, the CMA-AMLNN, CLPSO-AMLNN, and JADE-AMLNN algorithms demonstrated the highest performance with the same rank. Based on the chi-squared distribution table, with 0.05 degrees of freedom and a significance level of $\alpha=0.05$, the critical value is 21.03. Since the calculated chi-squared value is more significant than this critical value, the alternative hypothesis is accepted. This indicates a statistically significant difference between the algorithms. Additionally, the p-value is extremely small (1.2086E-08), further supporting the rejection of the null hypothesis ($H_{0}$).

The Wilcoxon signed-rank test is another non-parametric test employed in this paper, used to determine if there is a significant difference between the two algorithms.
% The resulting p-value indicates whether the median difference is significantly different from zero, which allows us to infer whether there is a significant change between the two algorithms. 
The Wilcoxon signed-rank test is selected over the t-test because the former does not assume normal distributions, making it a safer choice. Moreover, the Wilcoxon test is less influenced by outliers than the t-test\cite{tutorial_statistical}. Table~\ref{tab:wil} shows the results of the Wilcoxon signed-rank test with a significance level of 5\% based on accuracy. Based on the cumulative wins, we can see that CMA-AMLNN (7 wins, 4 ties, and 0 fail), JADE-AMLNN (8 wins, 4 ties, and 0 fail), and CLPSO-AMLNN (6 wins, 6 ties, and 0 fail) are the best-performing algorithms. On the other hand, GA-AMLNN (0 win, 1 ties, and 11 fails), SHADE-AMLNN (1 win, 6 ties, and 11 fails), and LSHADE-AMLNN (1 win, 0 ties, and 10 fails) are the worst-performing algorithms.

Despite providing valuable insights into the ranking and differences between the algorithms, both tests need more information regarding the algorithms' resistance to missing value. Consequently, a separate experiment was conducted to measure the algorithms' stability against the percentage of missing values, utilising the standard deviation criterion. In this context, a high standard deviation signifies that the algorithm is not resilient to missing value alterations, while a low standard deviation indicates increased resistance to missing value alterations. Figure~\ref{tab:std} shows the standard deviation, and it is clear that CMA-ES-AMLNN provides the lowest standard deviation, while JADE-AMLNN and CLPSO-AMLNN give higher standard deviations. These results are consistent with the previous tables. For instance, JADE-AMLNN's effectiveness is diminished when confronted with a substantial level of missing values (40\% in this case). As such, when robustness against missing value alteration is a crucial consideration, it can be concluded that CMA-ES-AMLNN stands out as the most resilient PBMH algorithm.

\section{Conclusion}
\label{sec:conc}

The decision-making process for smartphone purchases increasingly revolves around energy consumption considerations, driven by consumer awareness of sustainability issues. Despite the prevalence of energy-efficient practices in Android development, the lack of documented machine learning-based energy prediction algorithms tailored for mobile apps remains. To bridge this gap, our study introduces a novel neural network framework enhanced by metaheuristic techniques for robust energy prediction. Our metaheuristic approach proposed a novel approach to optimise learning algorithms and hyperparameters, as well as the architecture of neural networks. Furthermore, challenges related to incomplete data due to restricted access to certain device features are addressed through an algorithm selection strategy employing 13 metaheuristic algorithms. This strategy prioritises accuracy and resilience to missing values, ensuring the effectiveness of our predictive model.

%Energy usage becomes a vital determinant in the consumer's decision-making process when contemplating a smartphone purchase. Moreover, the importance of sustainability necessitates exploring approaches to reduce mobile device energy consumption, given the substantial global consequences stemming from the widespread use of smartphones, which profoundly impact the environment. Despite the presence of energy-efficient programming practices within the dominant Android platform, the scarcity of documented machine learning-based energy prediction algorithms specifically tailored for mobile app development remains evident. To address this gap, our research introduces a novel neural network-based framework, augmented by a metaheuristic approach, to achieve robust energy prediction. Our proposed metaheuristic approach plays a pivotal role in identifying suitable learning algorithms and their optimal parameters and determining the most appropriate number of layers and neurons in each layer. Additionally, the limitations in accessing certain aspects of a mobile phone may result in missing data in the data set. To address this challenge, we employed an optimal algorithm selection strategy, incorporating 13 metaheuristic algorithms, to identify the best algorithm based on accuracy and resistance to missing values. 

Notwithstanding the commendable performance demonstrated by the proposed approach, this work has opportunities for future extensions, which could be addressed by considering the following aspects:
\begin{itemize}
	\item The proposed approach employs a basic approach for handling missing values, while in the future, the performance can be enhanced by using some sophisticated missing value handling approaches. 
	\item It is important to note that new operating systems, their variants, and updates can influence the results of the proposed method. Therefore, the proposed algorithm could incorporate techniques such as continual learning to adapt online to these new updates.

\end{itemize}

\section{Acknowledgement}
This work was financed by FEDER (Fundo Europeu de Desenvolvimento Regional), from the European Union through CENTRO 2020 (Programa Operacional Regional do Centro), under project CENTRO-01-0247-FEDER-047256 – GreenStamp: Mobile Energy Efficiency Services.

This work is supported by NOVA LINCS (UIDB/04516/2020) with the financial support of FCT.IP.
 
\bibliography{jalal}

\begin{thebibliography}{10}
\providecommand{\url}[1]{#1}
\csname url@samestyle\endcsname
\providecommand{\newblock}{\relax}
\providecommand{\bibinfo}[2]{#2}
\providecommand{\BIBentrySTDinterwordspacing}{\spaceskip=0pt\relax}
\providecommand{\BIBentryALTinterwordstretchfactor}{4}
\providecommand{\BIBentryALTinterwordspacing}{\spaceskip=\fontdimen2\font plus
\BIBentryALTinterwordstretchfactor\fontdimen3\font minus
  \fontdimen4\font\relax}
\providecommand{\BIBforeignlanguage}[2]{{%
\expandafter\ifx\csname l@#1\endcsname\relax
\typeout{** WARNING: IEEEtran.bst: No hyphenation pattern has been}%
\typeout{** loaded for the language `#1'. Using the pattern for}%
\typeout{** the default language instead.}%
\else
\language=\csname l@#1\endcsname
\fi
#2}}
\providecommand{\BIBdecl}{\relax}
\BIBdecl

\bibitem{smartphone_03}
B.~Fu, J.~Lin, L.~Li, C.~Faloutsos, J.~Hong, and N.~Sadeh, ``Why people hate
  your app: Making sense of user feedback in a mobile app store,'' in
  \emph{Proceedings of the 19th ACM SIGKDD international conference on
  Knowledge discovery and data mining}, 2013, pp. 1276--1284.

\bibitem{smartphone_06}
I.~Manotas, C.~Bird, R.~Zhang, D.~Shepherd, C.~Jaspan, C.~Sadowski, L.~Pollock,
  and J.~Clause, ``An empirical study of practitioners' perspectives on green
  software engineering,'' in \emph{Proceedings of the 38th international
  conference on software engineering}, 2016, pp. 237--248.

\bibitem{smartphone_05}
G.~Pinto and F.~Castor, ``Energy efficiency: a new concern for application
  software developers,'' \emph{Communications of the ACM}, vol.~60, no.~12, pp.
  68--75, 2017.

\bibitem{Energy_APP08}
J.~Singh and A.~Maity, ``Energy consumption-based profiling of android apps,''
  in \emph{Mobile Application Development: Practice and Experience: 12th
  Industry Symposium in Conjunction with 18th ICDCIT 2022}.\hskip 1em plus
  0.5em minus 0.4em\relax Springer, 2023, pp. 21--32.

\bibitem{Energy_APP_09}
G.~Hecht, N.~Moha, and R.~Rouvoy, ``An empirical study of the performance
  impacts of android code smells,'' in \emph{Proceedings of the international
  conference on mobile software engineering and systems}, 2016, pp. 59--69.

\bibitem{Energy_APP_10}
S.~Mundody and K.~Sudarshan, ``Evaluating the impact of android best practices
  on energy consumption,'' in \emph{IJCA Proceedings on International
  Conference on Information and Communication Technologies}, vol.~8, 2014, pp.
  1--4.

\bibitem{Energy_APP07}
L.~Wattenbach, B.~Aslan, M.~M. Fiore, H.~Ding, R.~Verdecchia, and I.~Malavolta,
  ``Do you have the energy for this meeting? an empirical study on the energy
  consumption of the google meet and zoom android apps,'' in \emph{Proceedings
  of the 9th IEEE/ACM International Conference on Mobile Software Engineering
  and Systems}, 2022, pp. 6--16.

\bibitem{Energy_APP11}
Z.~Dai, W.~Wang, and Y.~Wu, ``Static energy consumption analysis for android
  applications,'' in \emph{IOP Conference Series: Earth and Environmental
  Science}, vol. 512, no.~1.\hskip 1em plus 0.5em minus 0.4em\relax IOP
  Publishing, 2020, p. 012011.

\bibitem{Energy_APP12}
J.~Singh and A.~Maity, ``Energy consumption-based profiling of android apps,''
  in \emph{Mobile Application Development: Practice and Experience: 12th
  Industry Symposium in Conjunction with 18th ICDCIT 2022}.\hskip 1em plus
  0.5em minus 0.4em\relax Springer, 2023, pp. 21--32.

\bibitem{smartphone_08}
M.~Nyman, ``Estimating the energy consumption of a mobile music streaming
  application using proxy metrics,'' Master's thesis, KTH, School of Electrical
  Engineering and Computer Science (EECS), 2020.

\bibitem{Energy_APP05}
E.~Iannone, M.~De~Stefano, F.~Pecorelli, and A.~De~Lucia, ``Predicting the
  energy consumption level of java classes in android apps: an exploratory
  analysis,'' in \emph{Proceedings of the 9th IEEE/ACM International Conference
  on Mobile Software Engineering and Systems}, 2022, pp. 1--5.

\bibitem{Energy_APP06}
A.~A. Bangash, K.~Ali, and A.~Hindle, ``A black box technique to reduce energy
  consumption of android apps,'' in \emph{Proceedings of the ACM/IEEE 44th
  International Conference on Software Engineering: New Ideas and Emerging
  Results}, 2022, pp. 1--5.

\bibitem{smartphone_09}
E.~Oliver and S.~Keshav, ``Data driven smartphone energy level prediction,''
  \emph{University of Waterloo Technical Report}, 2010.

\bibitem{smartphone_10}
C.~Luo, A.~Visuri, S.~Klakegg, N.~van Berkel, Z.~Sarsenbayeva,
  A.~M{\"o}tt{\"o}nen, J.~Goncalves, T.~Anagnostopoulos, D.~Ferreira, H.~Flores
  \emph{et~al.}, ``Energy-efficient prediction of smartphone unlocking,''
  \emph{Personal and Ubiquitous Computing}, vol.~23, pp. 159--177, 2019.

\bibitem{JPEG_Postdoc01}
S.~J. Mousavirad and L.~A. Alexandre, ``Energy-aware {JPEG} image compression:
  A multi-objective approach,'' \emph{Applied Soft Computing}, vol. 141, 2023.

\bibitem{JPEG_Postdoc02}
------, ``Metaheuristic-based energy-aware image compression for mobile app
  development,'' \emph{Multimedia Tools and Applications}, pp. 1--42, 2024.

\bibitem{JPEG_Postdoc03}
------, ``A metaheuristic-based machine learning approach for energy prediction
  in mobile app development,'' \emph{arXiv preprint arXiv:2306.09931}, 2023.

\bibitem{hyperparameter_optimization_survey}
L.~Yang and A.~Shami, ``On hyperparameter optimization of machine learning
  algorithms: Theory and practice,'' \emph{Neurocomputing}, vol. 415, pp.
  295--316, 2020.

\bibitem{GA_Main_Ref}
D.~Whitley, ``A genetic algorithm tutorial,'' \emph{Statistics and Computing},
  vol.~4, no.~2, pp. 65--85, 1994.

\bibitem{DE_Original}
R.~Storn and K.~Price, ``Differential evolution--a simple and efficient
  heuristic for global optimization over continuous spaces,'' \emph{Journal of
  Global Optimization}, vol.~11, no.~4, pp. 341--359, 1997.

\bibitem{Memetic_Main}
P.~Moscato \emph{et~al.}, ``On evolution, search, optimization, genetic
  algorithms and martial arts: Towards memetic algorithms,'' \emph{Caltech
  concurrent computation program, C3P Report}, vol. 826, p. 1989, 1989.

\bibitem{PSO_Main_Paper}
Y.~Shi and R.~Eberhart, ``A modified particle swarm optimizer,'' in \emph{IEEE
  International Conference on Evolutionary Computation}, 1998, pp. 69--73.

\bibitem{CMA_main_paper}
N.~Hansen and A.~Ostermeier, ``Completely derandomized self-adaptation in
  evolution strategies,'' \emph{Evolutionary Computation}, vol.~9, no.~2, pp.
  159--195, 2001.

\bibitem{HPSO-TVAC}
A.~Ratnaweera, S.~K. Halgamuge, and H.~C. Watson, ``Self-organizing
  hierarchical particle swarm optimizer with time-varying acceleration
  coefficients,'' \emph{IEEE Transactions on evolutionary computation}, vol.~8,
  no.~3, pp. 240--255, 2004.

\bibitem{Chaos-PSO}
B.~Liu, L.~Wang, Y.-H. Jin, F.~Tang, and D.-X. Huang, ``Improved particle swarm
  optimization combined with chaos,'' \emph{Chaos, Solitons \& Fractals},
  vol.~25, no.~5, pp. 1261--1271, 2005.

\bibitem{CLPSO02}
J.~J. Liang, A.~K. Qin, P.~N. Suganthan, and S.~Baskar, ``Comprehensive
  learning particle swarm optimizer for global optimization of multimodal
  functions,'' \emph{IEEE Transactions on Evolutionary Computation}, vol.~10,
  no.~3, pp. 281--295, 2006.

\bibitem{SAP_DE}
J.~Teo, ``Exploring dynamic self-adaptive populations in differential
  evolution,'' \emph{Soft Computing}, vol.~10, no.~8, pp. 673--686, 2006.

\bibitem{DE_Jade01}
J.~Zhang and A.~C. Sanderson, ``{JADE}: adaptive differential evolution with
  optional external archive,'' \emph{IEEE Transactions on Evolutionary
  Computation}, vol.~13, no.~5, pp. 945--958, 2009.

\bibitem{SHADE01}
R.~Tanabe and A.~Fukunaga, ``Success-history based parameter adaptation for
  differential evolution,'' in \emph{IEEE Congress on Evolutionary
  Computation}.\hskip 1em plus 0.5em minus 0.4em\relax IEEE, 2013, pp. 71--78.

\bibitem{LSHADE01}
R.~Tanabe and A.~S. Fukunaga, ``Improving the search performance of {SHADE}
  using linear population size reduction,'' in \emph{IEEE Congress on
  Evolutionary Computation}.\hskip 1em plus 0.5em minus 0.4em\relax IEEE, 2014,
  pp. 1658--1665.

\bibitem{Phasor-PSO}
M.~Ghasemi, E.~Akbari, A.~Rahimnejad, S.~E. Razavi, S.~Ghavidel, and L.~Li,
  ``Phasor particle swarm optimization: a simple and efficient variant of
  {PSO},'' \emph{Soft Computing}, vol.~23, no.~19, pp. 9701--9718, 2019.

\bibitem{ADAM-ADAMMAX}
D.~P. Kingma and J.~Ba, ``Adam: A method for stochastic optimization,''
  \emph{arXiv preprint arXiv:1412.6980}, 2014.

\bibitem{Adadelta}
M.~D. Zeiler, ``Adadelta: an adaptive learning rate method,'' \emph{arXiv
  preprint arXiv:1212.5701}, 2012.

\bibitem{ADAMW}
I.~Loshchilov and F.~Hutter, ``Decoupled weight decay regularization,''
  \emph{arXiv preprint arXiv:1711.05101}, 2017.

\bibitem{ASGD}
B.~T. Polyak and A.~B. Juditsky, ``Acceleration of stochastic approximation by
  averaging,'' \emph{SIAM journal on control and optimization}, vol.~30, no.~4,
  pp. 838--855, 1992.

\bibitem{NAdam}
T.~Dozat, ``Incorporating {Nesterov Momentum into Adam},'' in \emph{4th
  International Conference on Learning Representations}, 2016, pp. 1--4.

\bibitem{RAdam}
L.~Liu, H.~Jiang, P.~He, W.~Chen, X.~Liu, J.~Gao, and J.~Han, ``On the variance
  of the adaptive learning rate and beyond,'' \emph{arXiv preprint
  arXiv:1908.03265}, 2019.

\bibitem{RMSprop}
A.~Graves, ``Generating sequences with recurrent neural networks,'' \emph{arXiv
  preprint arXiv:1308.0850}, 2013.

\bibitem{RPROP}
M.~Riedmiller and H.~Braun, ``A direct adaptive method for faster
  backpropagation learning: The {RPROP} algorithm,'' in \emph{IEEE
  international conference on neural networks}.\hskip 1em plus 0.5em minus
  0.4em\relax IEEE, 1993, pp. 586--591.

\bibitem{SGD}
S.-i. Amari, ``Backpropagation and stochastic gradient descent method,''
  \emph{Neurocomputing}, vol.~5, no. 4-5, pp. 185--196, 1993.

\bibitem{greenhub_dataset}
R.~Pereira, H.~Matalonga, M.~Couto, F.~Castor, B.~Cabral, P.~Carvalho, S.~M.
  de~Sousa, and J.~P. Fernandes, ``Greenhub: a large-scale collaborative
  dataset to battery consumption analysis of android devices,'' \emph{Empirical
  Software Engineering}, vol.~26, pp. 1--55, 2021.

\bibitem{tutorial_statistical}
J.~Derrac, S.~Garc{\'\i}a, D.~Molina, and F.~Herrera, ``A practical tutorial on
  the use of nonparametric statistical tests as a methodology for comparing
  evolutionary and swarm intelligence algorithms,'' \emph{Swarm and
  Evolutionary Computation}, vol.~1, no.~1, pp. 3--18, 2011.

\end{thebibliography}
\bibliographystyle{IEEEtran}

\newpage

 %needed around the contents of the optional argument to biography to prevent
 %the LaTeX parser from getting confused when it sees the complicated
 %$\backslash${\tt{includegraphics}} command within an optional argument. (You can create
 %your own custom macro containing the $\backslash${\tt{includegraphics}} command to make things
 %simpler here.)
 
\vspace{11pt}

%\bf{If you include a photo:}\vspace{-33pt}
%\begin{IEEEbiography}[{\includegraphics[width=1in,height=1.25in,clip,keepaspectratio]{fig1}}]{Michael Shell}
%Use $\backslash${\tt{begin\{IEEEbiography\}}} and then for the 1st argument use $\backslash${\tt{includegraphics}} to declare and link the author photo.
%Use the author name as the 3rd argument followed by the biography text.
%\end{IEEEbiography}

\vspace{11pt}


\vfill

\end{document}